\pdfoutput=1

\documentclass[11pt]{article}

\usepackage{acl}

\usepackage{times}
\usepackage{latexsym}
\usepackage{amsmath}

\usepackage[T1]{fontenc}

\usepackage{graphicx}
\usepackage{booktabs}
\usepackage{tabularx}
\usepackage{tcolorbox}

\usepackage[utf8]{inputenc}

\usepackage{microtype}

\usepackage{inconsolata}
\usepackage{caption}
\usepackage{stfloats}
\usepackage{xspace}
\title{VIEScore: Towards Explainable Metrics for Conditional \\ Image Synthesis Evaluation}

\newcommand{\aspace}{\hspace{1em}}

\newcommand{\uwaterloo}{$^{\spadesuit}$}
\newcommand{\inai}{$^{\heartsuit}$}
\newcommand{\score}{\textsc{VIEScore}\xspace}

\author{
Max Ku\uwaterloo \aspace Dongfu Jiang\uwaterloo \aspace Cong Wei\uwaterloo \aspace Xiang Yue\inai \aspace Wenhu Chen\uwaterloo
\\
{\small{\texttt{\{m3ku, dongfu.jiang, c58wei\}@uwaterloo.ca, xiangyue@in.ai, wenhu.chen@uwaterloo.ca} }} \\ 
University of Waterloo\uwaterloo  \aspace
IN.AI Research\inai  \quad  \\
}

\begin{document}
\twocolumn[{%
    \renewcommand\twocolumn[1][]{#1}%
    \maketitle
    \centering
    \vspace{-10mm}
    \url{https://tiger-ai-lab.github.io/VIEScore/}
    \vspace{2mm}
    \begin{center}
        \centering
        \includegraphics[width=0.9\textwidth]{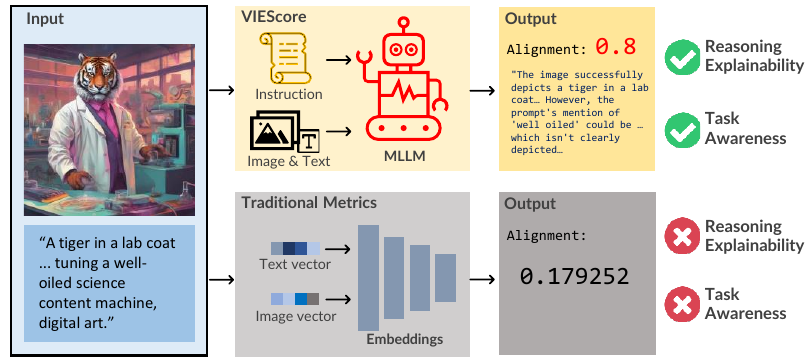}
        \captionof{figure}{Which method (\score or traditional metrics) is “closer” to the human perspectives? Metrics in the future would provide not just the score but also the rationale, enabling the understanding of each  judgment.}
        \label{fig:teaser}
    \end{center}
}]

\begin{abstract}
In the rapidly advancing field of conditional image generation research, challenges such as limited explainability lie in effectively evaluating the performance and capabilities of various models. This paper introduces \score, a Visual Instruction-guided Explainable metric for evaluating any conditional image generation tasks. \score leverages general knowledge from Multimodal Large Language Models (MLLMs) as the backbone and does not require training or fine-tuning. We evaluate \score on seven prominent tasks in conditional image tasks and found: (1) \score (GPT4-o) achieves a high Spearman correlation of 0.4 with human evaluations, while the human-to-human correlation is 0.45. (2) \score (with open-source MLLM) is significantly weaker than GPT-4o and GPT-4v in evaluating synthetic images. (3) \score achieves a correlation on par with human ratings in the generation tasks but struggles in editing tasks. With these results, we believe \score shows its great potential to replace human judges in evaluating image synthesis tasks.

\end{abstract}

\section{Introduction}
Diffusion models have become a focal point in AI research for image synthesis. Over the past year, several new models \citep{kumari2023multi, ruiz2023dreambooth, li2023dreamedit, zhang2023adding} have been introduced to enhance control over image generation. However, comprehensively evaluating AI-synthesized images remains a challenging and unresolved issue. While metrics like LPIPS \citep{zhang2018perceptual}, CLIP-Score \citep{hessel2021clipscore}, and DreamSim \citep{fu2023dreamsim} were proposed, they have certain limitations: (1) these metrics are agnostic the end task, which can fail to measure the desired aspects of the generated images, (2) the score is opaque with limited explainability. These limitations heavily restrict their effectiveness in assessing conditional image generation. Some research work~\citep{denton2015deep, isola2017image, meng2021sdedit, chen2023subject, sheynin2023emu} relied on human-driven evaluation methods. While humans excel at understanding and interpreting visual content, such methods in the context are facing challenges such as scalability limits and preference subjectivity issues. This reliance on human judgment highlights the need for more uniform evaluation methods in the field. To solve the mentioned issues, we formulate the problem definition with our desired properties, as presented in equation \ref{eq:VIE}. The function $f$ takes an instruction $I$, a synthesized image $O$, and $C^*$ which is a set of conditions (e.g. style, subject, background, canny-edge, etc). The score function should produce the intermediate rationale in the form of natural language before generating the final score according to the prompt instruction $I$:
\begin{equation}
    f_\text{VIE}(I, O, C^*) = (\text{rationale}, \text{score}) 
    \label{eq:VIE}
\end{equation}
The function $f$ can be any Multimodal Large Language Model (MLLM) such as GPT-4~\cite{openai2023gpt4} and LLaVA \citep{Liu2023VisualIT}, which can take input images to generate human-like text responses. Unlike the automatic metrics, MLLM can receive human instructions and produce rationale. With such motivation, we introduce \score (Visual Instruction-guided Explainable Score), a framework to assess synthetic images in different conditional image generation tasks. \score has multiple advantages compared to auto-metrics and human evaluation. It includes:

\noindent \textbf{Task Awareness.} Existing metrics were often designed to measure a certain aspect of generated images. For example, LPIPS measures the perceptual similarity of a pair of images, while CLIP-Score measures the text alignment of one single image. As a consequence, these metrics cannot be adapted to evaluate other tasks. \score acts as a silver bullet to tackle all conditional image generation evaluation processes due to its instruction-guiding property. It can be carefully adjusted with different instruction requirements.

\noindent \textbf{Explainability.} The existing metrics normally output a single float-point score, which cannot offer detailed insights into the 'rationale' behind its evaluations. Such a score makes it difficult to interpret the decisions from the metric output. Instead, \score can offer the rationale in the form of natural languages to help humans understand the reasoning process. As depicted in Figure \ref{fig:teaser}, the rationale can significantly improve the trustworthiness of \score.

\begin{figure}[!t]
    \centering
    \includegraphics[width=0.98\linewidth]{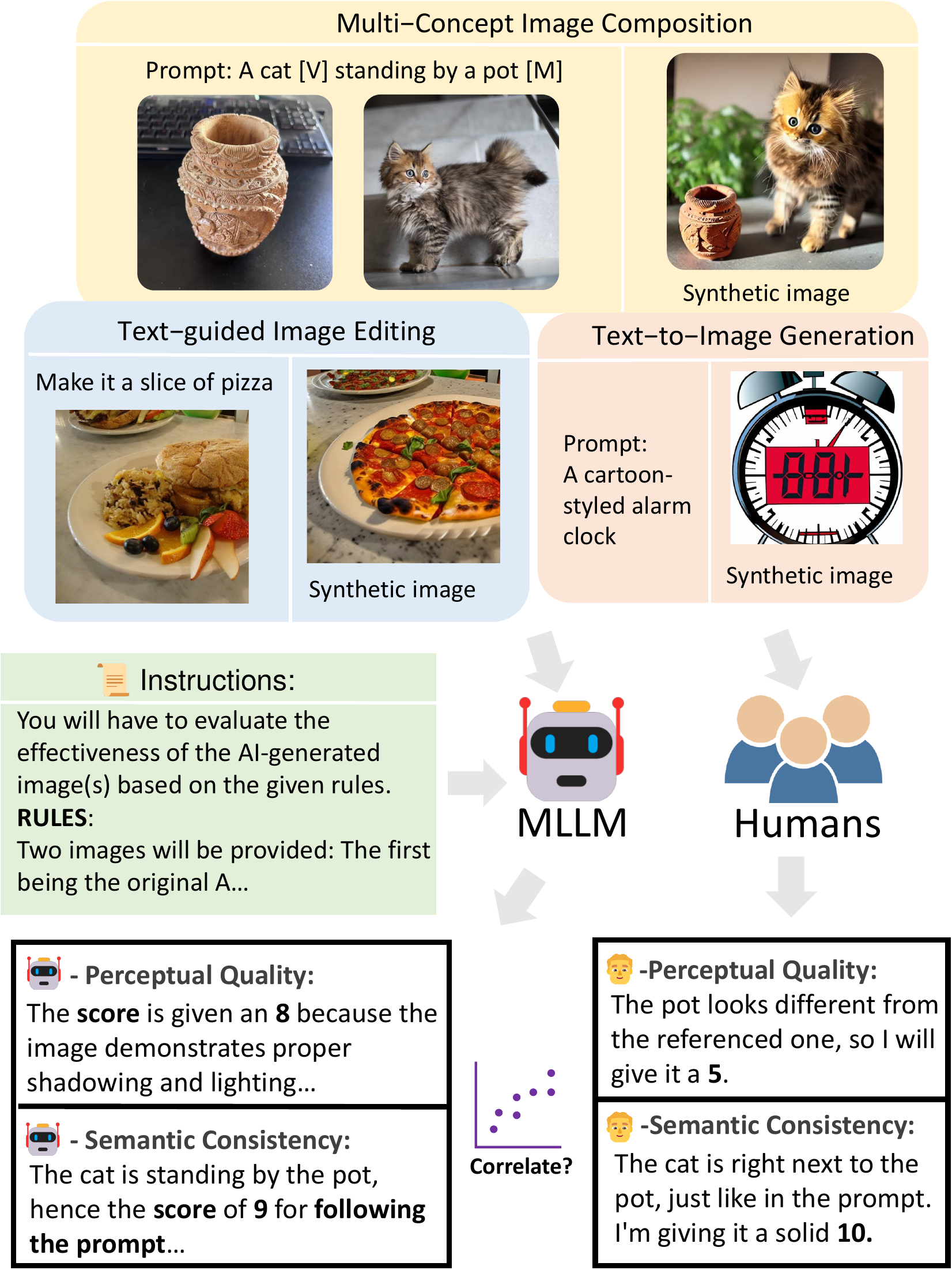}
    \caption{We study the correlation between MLLMs and human perspectives on rating images.}
    \vspace{-2ex}
    \label{fig:front_page}
\end{figure}

While the ultimate goal is to derive an MLLM that can rate images like humans, in this paper we also explore how well MLLMs can assess synthetic images compared to human evaluation and present insights and challenges on state-of-the-art MLLMs towards human evaluators, as shown in Figure \ref{fig:front_page}.

\section{Related Works}

\subsection{Conditional Image Synthesis}
With recent advancements in Image Synthesis research \citep{goodfellow2016deep, NEURIPS2020_4c5bcfec_diffusion, NEURIPS2021_49ad23d1_diffusion}, researchers proposed different methods and contributed a large amount of controllable image synthesis models with conditional inputs. Conditional image synthesis can be categorized into conditional image generation and conditional image editing tasks. Prevalent tasks include Text-To-Image generation \cite{saharia2022photorealistic, rombach2022high, sd-xl} (known as text-guided image generation), Inpainting \citep{avrahami2022blended, Lugmayr2022RePaintIU} (known as mask-guided image editing) and Text-guided image editing \citep{brooks2022instructpix2pix, couairon2022diffedit, cyclediffusion}. 

More recent works proposed new tasks such as Subject-driven image generation and editing \citep{gal2022image, ruiz2023dreambooth, li2023dreamedit} to inject one specific subject into a synthesized image, while Multi-concept image composition \citep{kumari2023multi, liu2023cones} allows multiple specific subjects into the synthesized image. Control-guided image generation
\citep{zhang2023adding, qin2023unicontrol} allows additional conditions alongside the text prompt to guide the image synthesis. Our work uses MLLM to access all the discussed tasks on synthetic image evaluation.

\subsection{Synthetic Images Evaluation}
Various metrics are proposed to evaluate the quality of AI-generated images. Traditional measures like the Inception Score (IS) \citep{salimans2016improved} and the Frechet Inception Distance (FID) \citep{heusel2017gans} are commonly employed to measure image fidelity. On the other hand, to measure the alignment between the generated image and the text prompt, several metrics \citep{kim2022mutual,kynkaanniemi2019improved,park2021benchmark,sajjadi2018assessing} have been introduced. The CLIP score \citep{hessel2021clipscore} and BLIP score \citep{li2022blip} are the most commonly used. Recently, approaches such as \citep{cho2023davidsonian} and \citep{lu2023llmscore} aim to provide a fine-grained evaluation framework, while the HEIM-benchmark \citep{lee2023holistic} assesses text-to-image models across multiple aspects, such as toxicity and safety. Other methods, such as projective-geometry \citep{Sarkar2023ShadowsDL}, evaluate images' physical and geometric realism. However, these metrics are primarily focused on text-to-image generation and remain narrow in scope. General image generation tasks like subject-driven image generation and image editing \citep{ruiz2023dreambooth, li2023dreamedit} still lack effective automatic metrics. One traditional, yet effective method to evaluate AI-generated image performance is to have human annotators assess visual quality. Recent works like ImagenHub \citep{ku2023imagenhub}, and HEIM~\cite{lee2023holistic} attempt to standardize human evaluation across various image generation tasks, though scalability remains a challenge. Our research aims to identify the challenges in mimicking human perception in synthetic image evaluation and address these gaps by developing auto-metrics that align with human judgment across common image evaluation tasks.

\subsection{Large Language Models as Evaluators} 

Large language models (LLMs) are often used to evaluate the quality of model-generated outputs. Recent works used LLMs as an evaluator demonstrating their great ability in text generation evaluation \citep{zheng2023judging,dubois2023alpacafarm}. This ability for evaluation naturally emerges \citep{Fu2023GPTScoreEA} and stems from LLM's great reasoning ability and instruction-following ability. Recent works also tried to devise a smaller but explicitly fine-tuned LLM\citep{Touvron2023LLaMAOA} that achieves similar evaluation results on natural language generation \citep{Xu2023INSTRUCTSCORETE,jiang2023TIGERScore,li2023generative}. Besides text evaluation, LLMs with visual features have been used as evaluators on images \citep{Lu2023LLMScoreUT, Huang2023T2ICompBenchAC, lee-etal-2020-vilbertscore, inan-etal-2021-cosmic-coherence, hu2023tifa}, mainly focused on evaluating text-to-image task. GPT-4v, regarded as the state-of-the-art LLM with visual features, also reported a decent ability on image evaluation, especially in text-image alignment \citep{Zhang2023GPT4visionAA}. However, the GPT-4v is not perfect for image evaluation. A comprehensive study on GPT-4v's vision ability reported that GPT-4v makes mistakes on image evaluation tasks \citep{yang2023dawn}. For example, it failed to provide proper reasonings for spotting the difference between two similar images.

\section{Preliminary}

\subsection{Evaluation Benchmark}
ImagenHub \citep{ku2023imagenhub} is a standardized benchmark for evaluating conditional image generation models with human raters. The framework covered mainstream tasks, including image generation, editing, and several conditioned tasks. In this section, we visit how humans assess images in the ImagenHub framework. Images are rated in two aspects: (1) Semantic Consistency (SC) assesses how well the generated image aligns with the given conditions, such as prompts and subject tokens, ensuring coherence and relevance to the specified criteria according to the task. (2) Perceptual Quality (PQ) evaluates the extent to which the generated image appears visually authentic and conveys a sense of naturalness. 

ImagenHub curated a human evaluation dataset for each task, in which the dataset contains around 100 to 200 conditional inputs for generating synthesized images. Then each image was rated by three human raters according to the guidelines of the defined task, and a final score in the range [0.0, 1.0] was reported for the average score in semantic consistency (SC) and perceptual quality (PQ) respectively, with another overall score (O) derived from the geometric mean of semantic consistency and perceptual quality at the instance level. ImagenHub covered 30 image synthesis models and reported 0.4 Krippendorff’s alpha on the inter-worker agreement of their human rating.

\subsection{Multimodal Large Language Models}
Multimodal Large Language Models (MLLMs) typically denote LLMs with integrated visual capabilities \citep{Yin2023ASO}. This visual proficiency opens up the potential to perform image analysis and evaluation. However, for a comprehensive assessment of synthetic images, multiple images may be examined in one pass due to complex conditions. The prompt will also be extensive to comprehensively describe the rating process. Therefore, the MLLM candidate should possess specific capabilities: (1) The model must efficiently process and interpret multiple images simultaneously. (2) The model needs to comprehend and respond to lengthy text prompts while matching all requirements. 

Recent popular open-source MLLMs, including LLaVA \citep{Liu2023VisualIT}, InstructBLIP \citep{Dai2023InstructBLIPTG}, Fuyu~\citep{fuyu-8b}, and  CogVLM~\citep{wang2023cogvlm}, can only accept a single image as input along with text instruction. To feed multiple images, a workaround is to merge and concatenate multiple images horizontally and feed as one image. More recent MLLMs such as Open-Flamingo~\citep{awadalla2023openflamingo}, Kosmos-2~\citep{Peng2023Kosmos2GM}, and QwenVL~\citep{Qwen-VL} can accept multiple images in an interleaved image-text format. For closed-source MLLMs, OpenAI's GPT-4v~\citep{openai2023gpt4}, GPT-4o~\citep{openai2023gpt4} and Google's Gemini~\citep{geminiteam2024gemini} are the most popular MLLMs that perform exceptionally.

\subsection{Existing Auto-metrics}
\label{sec:auto_metric_survey}
Here we list some prominent automatic metrics:

\noindent \textbf{Image-Text Alignment.} CLIP-Score \citep{hessel2021clipscore} computes the average cosine similarities between prompt and generated image CLIP embeddings. One disadvantage of CLIP-Score is that the score is biased towards the training distribution \citep{Kim2023BiastoTextDU}. Moreover, in practical evaluation, the average CLIP-Score result of a decent method will always fall in the range [0.25, 0.35] even though a single CLIP-Score is within [0, 1]. Such a narrow range may not offer enough differentiation to know which model is better. Moreover, image-text alignment is not the only considered aspect of semantic consistency. For example, it cannot examine the degree of overediting in text or mask-guided image editing tasks.\vspace{1ex}\\
\noindent\textbf{Perceptual Distance.} LPIPS \citep{zhang2018perceptual} measures the resemblance between two images in a manner that aligns with human perception. With its sensitivity to distortions, it is an often used metric in image synthesis research such as image editing tasks and control-guided Image Generation task \citep{meng2021sdedit, qin2023unicontrol}, to measure between the input (or ground truth) and the generated image. However, in the image editing context, the image's naturalness (e.g. shadow, lighting, sense of distance) is often required in the human perspective of perceptual quality, which is missed in the LPIPS metric. It is also difficult to access a model's performance by distortion level, as in the current state of research the models often can process high-quality editing without artifacts.\vspace{1ex}\\
\noindent\textbf{Subject Fidelity.} CLIP-I computes the average pairwise cosine similarities between CLIP embeddings of generated and real images, first proposed in Textual-Inversion \citep{gal2022image}. However, CLIP-I cannot distinguish between different subjects that may have highly similar text descriptions, and it is less sensitive to shape consistency as it compares the semantic similarity between images. DINO metric was proposed in DreamBooth \cite{ruiz2023dreambooth}. The metric is computed by the mean cosine similarities calculated pairwise between the DINO embeddings of ViT-S/16 \citep{Caron2021EmergingPI} for both synthesized and authentic images. In contrast to CLIP-I, the DINO metric is sensitive to differences between subjects of the same class due to the self-supervised training objective of DINO. These two became popular metrics reported in research on subject-driven image generation and editing tasks \citep{li2023dreamedit, Lu2023DreamComFT}.

\section{Method}

\begin{figure*}[!t]
    \centering
    \includegraphics[width=1\linewidth]{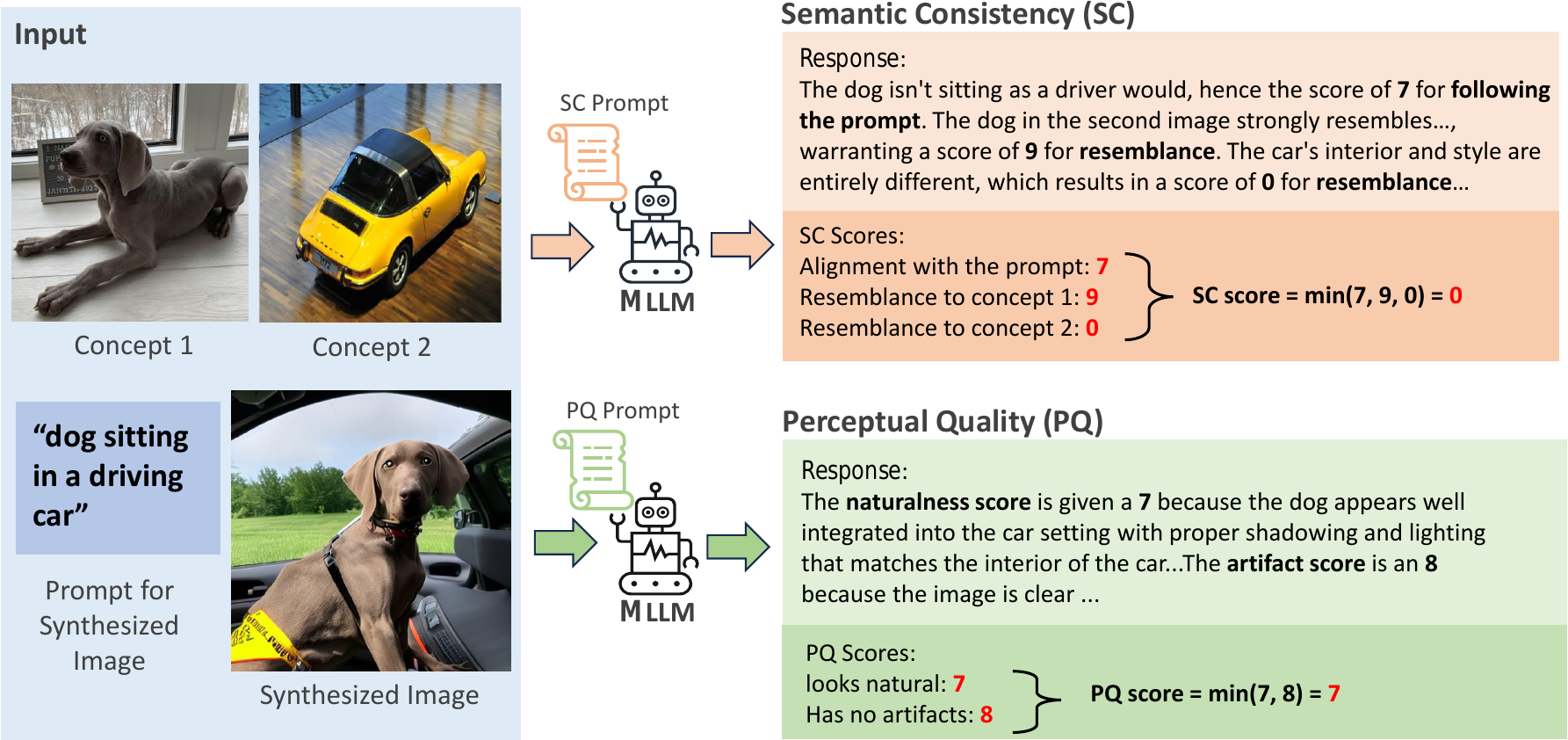}
    \caption{Process of MLLM evaluation on one synthetic image. All input conditions, synthesized images, and rating instructions are fed together to the MLLM in one pass. Multi-concept image composition task is used here as an example. The final overall score of the image is derived with equation \ref{eq:overall}.}
    \label{fig:method_main}
\end{figure*}

During the experiment, we select 29 models evaluated in ImagenHub \citep{ku2023imagenhub} to compare the correlations with human ratings. See Appendix \ref{sec:supp} for the listed models.

\noindent \textbf{Rating instructions.} In ImagenHub, each image in one rating aspect is rated by picking an option from List[0, 0.5, 1] by three human raters. While such simple rating instruction is human-friendly and offers enough granularity, the simplicity of the scale can lead to less accurate representations of opinions, as given the broad spectrum covered by the rating aspects of semantic consistency (SC) and perceptual quality (PQ). We propose a more rigorous rating instruction toward comprehensive evaluation for each type of task. We split the rating of semantic consistency (SC) and perceptual quality (PQ) into multiple sub-scores, which SC contains multiple scores according to the tasks. 

For example, in the multi-concept image composition task as shown in Figure \ref{fig:method_main}, two images (known as concepts) and a text prompt are provided as input, and the desired synthesized image will contain the two concept objects in the image in actions according to the text prompt. Thus SC will be split into 3 sub-scores: (1) Is the image aligning with the prompt? (2) Does the object in the image resemble the first concept? (3) Does the object in the image resemble the second concept? For PQ, the naturalness level and distortion level will be accessed separately, resulting in 2 sub-scores: (i) Does the image give an unnatural feeling such as a wrong sense of distance, wrong shadow, or wrong lighting? (ii) Does the image contain a portion of distortion, watermark, scratches, etc.? Our proposed rating system enhances the evaluation of tasks by dividing SC and PQ into distinct sub-scores. The details of prompt templates are available in Appendix \ref{sec:appendix_prompt}.
\begin{equation}
    O = [\min(\alpha_1, ..., \alpha_i) \min(\beta_1, ..., \beta_i)]^{\frac{1}{2}}
    \label{eq:overall}
\end{equation}
Our overall score is derived as shown in equation \ref{eq:overall}. We assume each sub-score weights the same and used $\min$ operation to emphasize the importance of meeting all criteria without exception. $\alpha_i$ is a sub-score in SC and $\beta_i$ is a sub-score in PQ. The final rating scores of SC and PQ provided by MLLMs are on a scale of 0 to 10. The design rationale is that in ImagenHub's human rating method, the possible results when the answers of three human raters, each picking an option from List[0, 0.5, 1], are added together and then divided by 3, will fall into one of the options: List[0.0, 0.17, 0.33, 0.5, 0.67, 0.83, 1.0]. Thus we simply use a scale of 0 to 10 and normalized in the range [0.0, 1.0] when comparing with human ratings. Input conditions and synthetic image are fed into the MLLM together during the rating process of SC, while in the PQ rating process, only the synthetic image is fed into the MLLM. This is to avoid the model getting confused by the input conditions in the PQ rating process, as to be discussed in section \ref{sec:corr_study}.

\section{Experimental Results}

\begin{table}[!t]
\centering
\scalebox{0.9}{
\begin{tabular}{l|ccc}
\toprule
Backbone & M-H$^{SC}_{\text{corr}}$ & M-H$^{PQ}_{\text{corr}}$ & M-H$^{O}_{\text{corr}}$ \\
\midrule
\multicolumn{4}{c}{\textbf{Across All 7 Tasks}} \\
\midrule
Human Raters & 0.4700 & 0.4124 & 0.4558 \\
\midrule
\multicolumn{4}{c}{\score} \\
GPT-4o$_\text{0shot}$ & \colorbox{green}{0.4459} &  \colorbox{green}{0.3399} & \colorbox{green}{0.4041} \\
GPT-4o$_\text{1shot}$ &  0.4309 &  0.1167 &  0.3770 \\
Gemini-Pro$_\text{0shot}$ & 0.3322 &  0.2675 & 0.3048 \\
Gemini-Pro$_\text{1shot}$ & 0.3094 &  0.3070 & 0.3005 \\
GPT-4v$_\text{0shot}$ & 0.3655 &  0.3092 & 0.3266 \\
GPT-4v$_\text{1shot}$ &  0.2689 &  0.2338 &  0.2604 \\
LLaVA$_\text{0shot}$ &  0.1046 &  0.0319 &  0.0925 \\
LLaVA$_\text{1shot}$ & 0.1012 & 0.0138 & 0.0695 \\
Qwen-VL$_\text{0shot}$         & 0.0679 & 0.0165 & 0.0920 \\
BLIP2$_\text{0shot}$           & 0.0504 & -0.0108 & 0.0622 \\
InstructBLIP$_\text{0shot}$    & 0.0246 & 0.0095 & 0.0005 \\
Fuyu$_\text{0shot}$            & -0.0110 & -0.0172 & 0.0154 \\
CogVLM$_\text{0shot}$          & -0.0228 & 0.0514 & -0.0050 \\
OpenFlamingo$_\text{0shot}$    & -0.0037 & -0.0102 & -0.0122 \\
\bottomrule
\end{tabular} }
\caption{Correlations across all tasks with different backbone models. We highlight the highest correlation numbers in green. See Appendix \ref{sec:backbone_performance} for details.}
\label{tab:result_all}
\end{table}

\begin{table}[!t]
\centering
\scalebox{0.9}{
\begin{tabular}{l|ccc}
\toprule
Method & M-H$^{SC}_{\text{corr}}$ & M-H$^{PQ}_{\text{corr}}$ & M-H$^{O}_{\text{corr}}$ \\
\midrule
\multicolumn{4}{c}{\textbf{Text-guided Image Generation} (5 models)} \\
\midrule
Human Raters & 0.5044 & 0.3640 & 0.4652 \\
CLIP-Score & -0.0817 & -0.0114 & -0.0881 \\
\midrule
\multicolumn{4}{c}{\score} \\
GPT-4o$_\text{0shot}$ & \colorbox{green}{0.4989} & \colorbox{green}{0.2495} & 0.3928 \\
GPT-4o$_\text{1shot}$ &  0.5124 &  0.0336 &  0.4042 \\
Gemini-Pro$_\text{0shot}$ & 0.5123 &  0.1842 & 0.4356 \\
Gemini-Pro$_\text{1shot}$ & 0.4757 &  0.2206 & 0.4326 \\
\colorbox{yellow}{GPT-4v$_\text{0shot}$} & 0.4885 &  0.2379 & \colorbox{green}{0.4614} \\
GPT-4v$_\text{1shot}$ & 0.4531 & 0.1770 & 0.3801 \\
LLaVA$_\text{0shot}$ &  0.1809 &  0.0306 &  0.1410 \\
LLaVA$_\text{1shot}$ & 0.1789 & -0.0020 & 0.1309 \\
\midrule
\multicolumn{4}{c}{\textbf{Mask-guided Image Editing} (4 models)} \\
\midrule
Human Raters & 0.5390 & 0.5030 & 0.4981 \\
LPIPS & -0.1012 & 0.0646 & -0.0694 \\
\midrule
\multicolumn{4}{c}{\score} \\
\colorbox{yellow}{GPT-4o$_\text{0shot}$} & \colorbox{green}{0.5421} & \colorbox{green}{0.3469} & \colorbox{green}{0.4769} \\
GPT-4o$_\text{1shot}$ &  0.5246 &  0.1272 &  0.4432 \\
Gemini-Pro$_\text{0shot}$ & 0.4304 &  0.2839 & 0.3593 \\
Gemini-Pro$_\text{1shot}$ & 0.4595 &  0.3170 & 0.4017 \\
GPT-4v$_\text{0shot}$ &  0.4508 &  0.2859 & 0.4069 \\
GPT-4v$_\text{1shot}$ & 0.4088 & 0.2352 & 0.3810 \\
LLaVA$_\text{0shot}$ &  0.1180 &  -0.0531 &  0.0675 \\
LLaVA$_\text{1shot}$ & 0.1263 & -0.0145 & 0.1040 \\
\midrule
\multicolumn{4}{c}{\textbf{Text-guided Image Editing} (8 models)} \\
\midrule
Human Raters & 0.4230 & 0.5052 & 0.4184 \\
LPIPS & 0.0956 & 0.2504 & 0.1142 \\
\midrule
\multicolumn{4}{c}{\score} \\
\colorbox{yellow}{GPT-4o$_\text{0shot}$} & \colorbox{green}{0.4062} & \colorbox{green}{0.4863} & \colorbox{green}{0.3821} \\
GPT-4o$_\text{1shot}$ &  0.3684 &  0.1939 &  0.3438 \\
Gemini-Pro$_\text{0shot}$ & 0.2836 &  0.4291 & 0.2728 \\
Gemini-Pro$_\text{1shot}$ & 0.2805 &  0.4657 & 0.2648 \\
GPT-4v$_\text{0shot}$ &  0.2610 & 0.4274 &  0.2456 \\
GPT-4v$_\text{1shot}$ & 0.2428 & 0.3402 & 0.2279 \\
LLaVA$_\text{0shot}$ & 0.0448 & 0.0583 &  0.0273 \\
LLaVA$_\text{1shot}$ & 0.0185 & -0.0107 & 0.0258 \\
\bottomrule
\end{tabular} }
\caption{Correlations comparison of available methods on the most common tasks. We highlight the best method and the correlation numbers closest to human raters. Continue in Table~\ref{tab:result_main_2} and \ref{tab:result_main_3}.}
\label{tab:result_main_1}
\end{table}

\begin{table}[!t]
\centering
\scalebox{0.9}{
\begin{tabular}{l|ccc}
\toprule
Method & M-H$^{SC}_{\text{corr}}$ & M-H$^{PQ}_{\text{corr}}$ & M-H$^{O}_{\text{corr}}$ \\
\midrule
\multicolumn{4}{c}{\textbf{Subject-driven Image Generation} (4 models)} \\
\midrule
Human Raters & 0.4780 & 0.3565 & 0.4653 \\
DINO & 0.4160 & 0.1206 & 0.4246 \\
CLIP-I & 0.2961 & 0.1694 & 0.3058 \\
\midrule
\multicolumn{4}{c}{\score} \\
\colorbox{yellow}{GPT-4o$_\text{0shot}$} & \colorbox{green}{0.4806} &  \colorbox{green}{0.2576} & \colorbox{green}{0.4637} \\
GPT-4o$_\text{1shot}$ &  0.4685 &  -0.0171 &  0.4292 \\
Gemini-Pro$_\text{0shot}$ & 0.2906 &  0.1765 & 0.2851 \\
Gemini-Pro$_\text{1shot}$ & 0.3486 &  0.2800 & 0.3342 \\
GPT-4v$_\text{0shot}$ & 0.3979 &  0.1903 & 0.3738 \\
GPT-4v$_\text{1shot}$ & 0.2757 & 0.2261 & 0.2753 \\
LLaVA$_\text{0shot}$ &  0.0326 &  -0.0303 &  0.1219 \\
LLaVA$_\text{1shot}$ & 0.1334 & 0.0858 & 0.1248 \\
\midrule
\multicolumn{4}{c}{\textbf{Subject-driven Image Editing} (3 models)} \\
\midrule
Human Raters & 0.4887 & 0.2986 & 0.4747 \\
DINO & 0.3022 & -0.0381 & 0.3005 \\
CLIP-I & 0.2834 & 0.1248 & 0.2813 \\
\midrule
\multicolumn{4}{c}{\score} \\
\colorbox{yellow}{GPT-4o$_\text{0shot}$} & \colorbox{green}{0.4800} & \colorbox{green}{0.3734} & \colorbox{green}{0.3268} \\
GPT-4o$_\text{1shot}$ &  0.3862 &  0.1273 &  0.2797 \\
Gemini-Pro$_\text{0shot}$ & 0.2187 &  0.3148 & 0.2234 \\
Gemini-Pro$_\text{1shot}$ & -0.0083 &  0.3181 & 0.0004 \\
GPT-4v$_\text{0shot}$& 0.3274 &  0.2960 &  0.1507 \\
GPT-4v$_\text{1shot}$ & -0.0255 & 0.1572 & -0.0139 \\
LLaVA$_\text{0shot}$ &  0.0360 &  -0.0073 &  0.0168 \\
LLaVA$_\text{1shot}$ & 0.0587 &  -0.0249 & 0.0309 \\
\bottomrule
\end{tabular} }
\caption{Correlations comparison of available methods on the subject-driven tasks.}
\label{tab:result_main_2}
\end{table}

\begin{table}[!t]
\centering
\scalebox{0.9}{
\begin{tabular}{l|ccc}
\toprule
Method & M-H$^{SC}_{\text{corr}}$ & M-H$^{PQ}_{\text{corr}}$ & M-H$^{O}_{\text{corr}}$ \\
\midrule
\multicolumn{4}{c}{\textbf{Multi-concept Image Composition} (3 models)} \\
\midrule
Human Raters & 0.5927 & 0.5145 & 0.5919 \\
DINO & 0.0979 & -0.1643 & 0.0958 \\
CLIP-I & 0.1512 & -0.0963 & 0.1498 \\
\midrule
\multicolumn{4}{c}{\score} \\
\colorbox{yellow}{GPT-4o$_\text{0shot}$} & \colorbox{green}{0.4516} & 0.2751 & \colorbox{green}{0.4136} \\
GPT-4o$_\text{1shot}$ &  0.4120 &  -0.0141 &  0.3523 \\
Gemini-Pro$_\text{0shot}$ & 0.3557 &  0.1948 & 0.3314 \\
Gemini-Pro$_\text{1shot}$ & 0.4151 &  0.1798 & 0.4131 \\
GPT-4v$_\text{0shot}$ & 0.3209 & \colorbox{green}{0.3025} & 0.3346 \\
GPT-4v$_\text{1shot}$ & 0.1859 & 0.1185 & 0.1918 \\
LLaVA$_\text{0shot}$ &  0.1022 &  0.1194 &  0.1070 \\
LLaVA$_\text{1shot}$ & 0.0828 & 0.0379 & 0.0293 \\
\midrule
\multicolumn{4}{c}{\textbf{Control-guided Image Generation} (2 models)} \\
\midrule
Human Raters & 0.5443 & 0.5279 & 0.5307 \\
LPIPS & 0.3699 & 0.4204 & 0.4133 \\
\midrule
\multicolumn{4}{c}{\score} \\
\colorbox{yellow}{GPT-4o$_\text{0shot}$} & \colorbox{green}{0.4972} & \colorbox{green}{0.4892} & \colorbox{green}{0.5439} \\
GPT-4o$_\text{1shot}$ &  0.5544 &  0.3699 &  0.5238 \\
Gemini-Pro$_\text{0shot}$ & 0.3254 &  0.3359 & 0.2960 \\
Gemini-Pro$_\text{1shot}$ & 0.2677 &  0.4392 & 0.3240 \\
GPT-4v$_\text{0shot}$ & 0.4360 & 0.4975 & 0.3999 \\
GPT-4v$_\text{1shot}$ & 0.3892 & 0.4132 & 0.4237 \\
LLaVA$_\text{0shot}$ &  0.2207 &  0.1060 &  0.1679 \\
LLaVA$_\text{1shot}$ & 0.1121 & 0.0247 & 0.0416 \\
\bottomrule
\end{tabular} }
\caption{Correlations comparison of available methods on the control-guided and multi-concept tasks.}
\label{tab:result_main_3}
\end{table}

\subsection{Correlation Study}
\label{sec:corr_study}

For all presented correlations, we applied Fisher Z-transformation to estimate the average Spearman correlation $\in [-1, 1]$ across models and tasks. 

\noindent \textbf{Metric-to-Human (M-H) correlations.} In Table \ref{tab:result_main_1} and \ref{tab:result_main_2}, we first verified the reliability of ImagenHub human ratings by computing the Human-to-Human (H-H) correlation, as the correlation goes around 0.5, expected to be the highest value compared to MLLMs. Then we benchmark the MLLMs according to our designed method to compute the Metric-to-Human (M-H) correlation. We noticed only GPT-4v, GPT4o, Gemini, and LLaVA were able to follow our instructions clearly while other MLLMs were not able to produce any meaningful results according to our setup. For example, BLIP-2, while able to output the correct format, the scores provided are constant zeros. Qwen-VL and InstructBLIP could only produce a portion of responses for semantic consistency but failed to generate any results for perceptual quality evaluation. From overall performance, we found that GPT-4o reports a significantly higher correlation than all other MLLMs, while LLaVA's correlation is much less than human raters. It seems that LLaVA is less effective in these specific tasks compared to close-sourced MLLMs like GPT-4v. It is worth mentioning that GPT-4o achieves a very high correlation with human raters on different image generation and editing tasks. GPT-4v and Gemini also show satisfactory performance on nearly all tasks with a difference of less than 0.2 towards human correlations, even on par with humans in text-guide image generation tasks. Both GPT-4v, Gemini, and LLaVA demonstrated the weakest performance in the text-guided editing task and subject-driven image editing task.

\noindent \textbf{Extra visuals resulted in a decline in performance.} Numerous studies \citep{Brown2020LanguageMA, Parnami2022LearningFF, Liu2021WhatMG} have highlighted that In-Context Learning (ICL) allows LLMs to tackle novel tasks effectively without requiring the traditional fine-tuning process. We applied In-Context Learning in our prompting method with the expectation of increasing the correlation scores, but we observed the opposite. In Table \ref{tab:result_main_1} and \ref{tab:result_main_2}, there is an observable general trend of diminishing correlation scores. The overall correlation score in subject-driven image generation and editing, and the multi-concept image composition task dropped significantly. Only the mask-guided image editing task and control-guided image generation task reported a subtle increase in correlation score.

\begin{figure}[!t]
    \centering
     \includegraphics[width=1\linewidth]{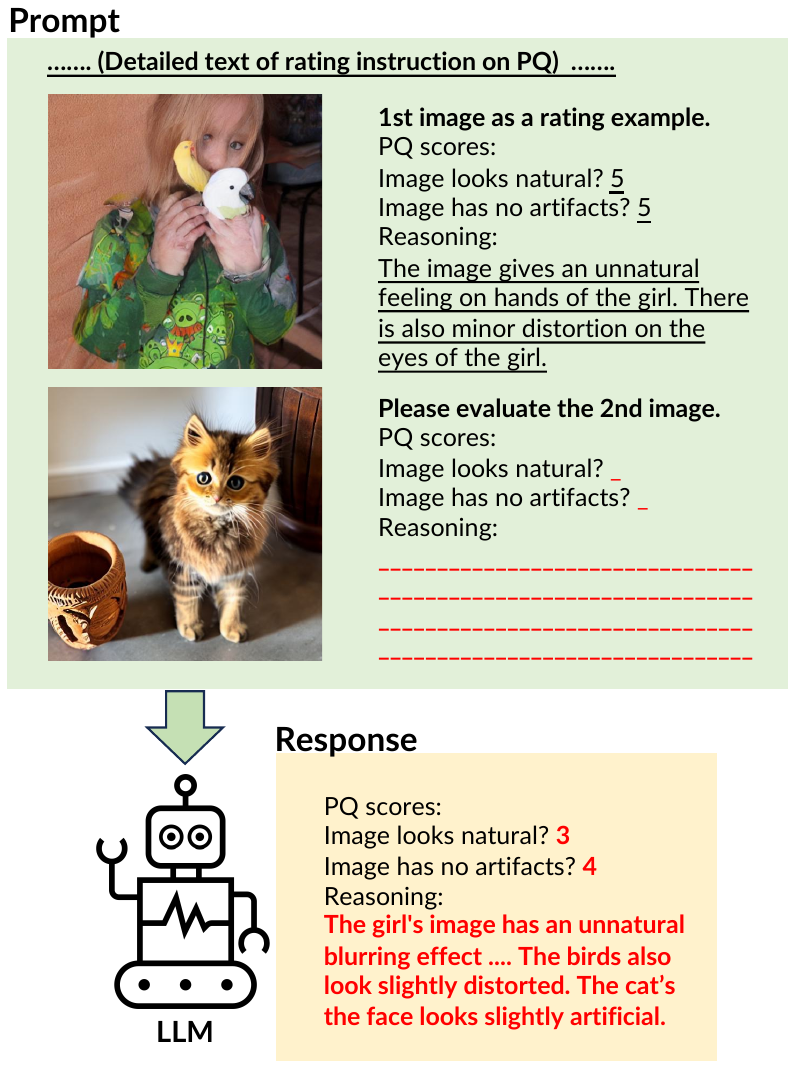}
    \caption{MLLM making mistakes on rationale when prompted with extra images as examples.}
    \vspace{-2ex}
    \label{fig:PQ_1shot}
\end{figure}

Looking into the rationale, we found that the MLLMs tend to get confused by the example images, as illustrated in Figure \ref{fig:PQ_1shot}. Such behavior is observed in both GPT-4v, GPT-4o, Gemini, and LLaVA rationale. Another recent work \citep{Lu2023VIMPM} also reported a similar issue where the model attempted to consider the example when answering the visual question. This explains the deterioration of the correlation on both GPT-4v, GPT-4o, Gemini, and LLaVA when the ICL prompting technique is used. This also implied the low correlation scores on image editing tasks were due to the limited capability of state-of-the-art MLLMs for multiple image understanding.

\noindent \textbf{Ablation study on PQ rating setting.}
As providing multiple images could potentially decrease the performance, we attempt to minimize the workload of MLLM by only providing the synthetic image in the PQ rating process instead of including the input conditions. We report the correlation score in the two different settings with GPT-4v in Table \ref{tab:PQ_ablation} to examine the impact. We spotted a significant improvement in correlation after taking away the inputs in the PQ rating process.

\begin{table}[!t]
\centering
\scalebox{0.9}{
\begin{tabular}{l|cc}
\toprule
& TIE & MCIC \\
PQ Prompting Method & M-H$^{PQ}_{\text{corr}}$ & M-H$^{PQ}_{\text{corr}}$ \\
\midrule
Human (with inputs) & 0.5052 & 0.5145 \\
\midrule
without inputs & 0.4274 &  0.3025 \\
with inputs &  0.2256 &  0.0731 \\
\bottomrule
\end{tabular} }
\caption{Correlations of GPT-4v when including inputs in the PQ prompt in TIE (Text-guided Image Editing) and MCIC (Multi-concept Image Composition) task. See \ref{tab:app_GPT-4v_human_corr} for detailed comparison in Appendix.}
\label{tab:PQ_ablation}
\end{table}

\noindent \textbf{Ranking image models.} Besides rating score correlations, we also compared the model ranking from the ImagenHub human evaluation leaderboard and the model ranking suggested by the MLLMs, shown in Table \ref{tab:ranking_corr}. We computed Spearman's footrule $d_{SF}(r,r_*)\in[0,+\infty)$ and Spearman's rho $\rho_{S}(r,r_*)\in[-1,1]$ to examine the ranking correlation. Both GPT-4v and LLaVA can align to ImagenHub rankings on the multi-concept image composition task and control-guided image generation task, and with only one model difference in the subject-driven image editing task. While the results vary significantly across other tasks, GPT-4v generally maintains a stronger alignment with the ImagenHub rankings compared to LLaVA.

\begin{table*}[!t]
\centering
\scalebox{0.85}{
\begin{tabular}{l|cccc|cccc}
\toprule
& \multicolumn{4}{c}{$d_{SF}(r_\text{Human}, r_{\text{Method}})$↓}  & \multicolumn{4}{c}{$\rho_{S}(r_\text{Human}, r_{\text{Method}})$↑}  \\
Task (Total number of Models) & GPT-4v & LLaVA & LPIPS & CLIP & GPT-4v & LLaVA & LPIPS & CLIP \\
\midrule
Text-guided Image Generation (5) & 2 & 6 & N/A & 8 & 0.90 & 0.50 & N/A & -0.20 \\
Mask-guided Image Editing (4) & 2 & 8 & 2 & 0 & 0.80 & -1.00 & 0.80 & 1.00\\
Text-guided Image Editing (8) & 12 & 16 & 20 & 16 & 0.67 & 0.48 & 0.17 & 0.48\\
Subject-driven Image Generation (4) & 4 & 6 & 0 & 6 & 0.20 & -0.40 & 1.00 & -0.20\\
Subject-driven Image Editing (3) & 2 & 2 & 4 & 4 & 0.50 & 0.50 & -0.50 & -1.00\\
Multi-concept Image Composition (3) & 0 & 0 & 2 & 2 & 1.00 & 1.00 & 0.50 & 0.50\\
Control-guided Image Generation (2) & 0 & 0 & 0 & 2 & 1.00 & 1.00 & 1.00 & -1.00\\
\bottomrule
\end{tabular} }
\caption{Ranking judgment from each metric method. $d_{SF}(r,r_*)$ is the Spearman's footrule and $\rho_{S}(r,r_*)$ is the Spearman's rho (correlation). LPIPS is not available for the first task because there are no reference images.}
\label{tab:ranking_corr}
\end{table*}

\subsection{Insights and Challenges on \score}

\begin{figure}[!t]
    \centering
    \includegraphics[width=1\linewidth]{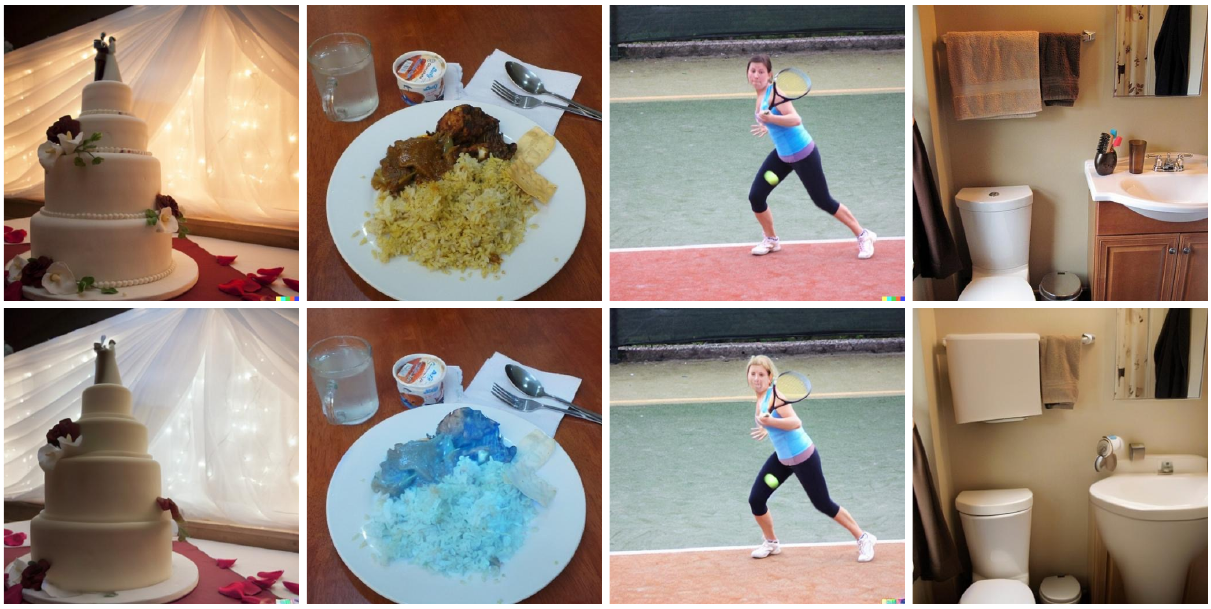}
    \caption{Representative pairs that MLLMs misunderstood as identical images.
    Images in the first row are the inputs and in the second row are the edited.}
    \vspace{-2ex}
    \label{fig:ID_0shot}
\end{figure}

\textbf{MLLMs are weak at capturing image nuances in edited images.} 
From Table \ref{tab:result_main_1}, we noticed the correlation scores on editing tasks are generally lower than generation tasks. Upon investigation, it was found that MLLMs often fail to detect minor changes made in image editing, such as small patch edits. Consequently, MLLMs might perceive two images as identical even when humans recognize the edits as successful. This issue may stem from MLLMs focusing on high-level image features while overlooking finer details like color and texture differences, as illustrated in Figure \ref{fig:ID_0shot}. This limitation is apparent in both GPT-4v, GPT-4o, Gemini, and LLaVA, highlighting a challenge in synthetic image evaluation accuracy on image editing tasks. 

\noindent \textbf{Both MLLMs and human evaluators display a broader range of views regarding perceptual quality compared to semantic consistency.} From Table \ref{tab:result_main_1} and Table \ref{tab:result_main_2}, we can observe that the correlation scores of PQ are generally lower than the correlation scores of SC and Overall, even on human raters. This suggests the human raters' perspective on evaluating perceptual quality is more diverse. Possible impacting factors include the rater’s eyesight condition, screen resolution, rating leniency, etc. In the context of MLLMs, we found that MLLMs while being able to correctly recognize the naturalness and artifacts of the image, the rating scores are as diverse as human rating scores even though we have provided a marking rubric.


\subsection{\score and Auto-metrics vs Human}

We report the human correlations in Table \ref{tab:result_main_1} and \ref{tab:result_main_2} to compare the performance between our \score and popular auto-metrics. To ensure a fair comparison, we only included automatic metrics that have been previously reported in related research for the specific tasks under consideration.

\noindent \textbf{DINO is an effective metric in subject-driven tasks.} The DINO metric demonstrates sensitivity to variations within the same class of subjects, making it an effective metric for measuring whether the subject in the synthesized image aligns with the token subject. Our correlation result shows that DINO outperforms GPT-4v and CLIP-I on subject-driven image generation and editing tasks, suggesting that DINO highly aligns with human's perspective on semantic consistency where subject fidelity is considered. However, multi-subject fidelity remains a challenge as DINO and CLIP-I only consider single subjects. However, GPT-4o still achieves the highest correlation to human annotations.

\noindent \textbf{LPIPS metric proves to be effective in control-guided tasks, but less effective in image editing tasks.} As discussed in section \ref{sec:auto_metric_survey}, LPIPS has great ability in detecting distortions. Since the control-guided task is a less mature research direction compared to image editing tasks, distortions are often found in the synthetic images from the control-guided task. On the other hand, current image editing models can synthesize images with less distortions. This explains the high correlation in the control-guided task. While GPT-4o still achieves the highest correlation to human annotations, LPIPS outperforms GPT-4v, Gemini, and LLaVA on this task. 

\noindent \textbf{CLIP-Score has a much weaker correlation with human ratings in the text-guided image generation task than GPT-4v.} We also noticed none of the synthetic images achieved higher than 0.3 CLIP-Score, even though most of the images are regarded as having high semantic consistency by human raters. This can be due to different evaluation focuses, as humans tend to grab the abstract idea from the prompt to access the image, but CLIP-Score considers the whole text prompt.

\noindent \textbf{GPT-4v outperforms other auto-metrics with its correlation to the ImagenHub leaderboard rankings.} The correlation of model rankings on ImagenHub was evaluated against CLIP Score and LPIPS metrics, as shown in Table \ref{tab:ranking_corr}, and compared with MLLMs in the \score. We found that GPT-4v can achieve a positive correlation with the model rankings on every task. This shows the sign of capability for MLLMs as evaluators for image synthesis research.

\section{Conclusion}

In this paper, we propose the \score for synthetic image evaluation across seven popular image synthesis tasks and comprehensively access the efficacy using human ratings from ImagenHub. Our experiment reported that \score with close-source  MLLMs backbones like GPT-4o and GPT-4v are significantly more effective than other open-source MLLMs in assessing synthetic images, achieving a correlation of over 0.4 to human ratings most of the tasks. However, it notes a lower correlation in image editing tasks for most of the MLLMs, including GPT-4v. We also noticed Gemini has similar performance as GPT-4v, while GPT-4o stands superior. Comparing our \score to existing auto-metrics, we found that GPT-4o is more effective than auto-metrics in all tasks, while DINO is more effective in subject-driven image generation and editing tasks than GPT-4v. GPT-4v also shows a higher ranking correlation with the ImagenHub leaderboard than other automatic metrics. This marked a milestone towards explainable metrics for conditional image synthesis evaluation. Our future research will focus on investigating the use of distillation models to replicate human-like performance in evaluating synthetic images.

\section{Limitations}
\noindent \textbf{OpenAI Security and Privacy Policy.} Due to ChatGPT's security and privacy policy, AI-generated images that resemble a real person or photograph will be refused by GPT-4v for evaluation. The model will return results similar to "I am sorry, but I cannot process these images as they contain real people.". We simply drop those results by keyword matching.

\noindent \textbf{OpenAI Playground vs API.} While GPT-4v Playground allows the user to keep a session, the OpenAI API does not provide such a function. While we believe using GPT-4v playground might yield better performance, especially in an In-Context learning setting, we can only rely on API due to the large scale of the experiment. 

\section{Potential Risks}
Multimodal models can inadvertently perpetuate or amplify biases present in their training data. The interpretation and evaluation of synthetic images depend heavily on context. A multimodal model might not fully grasp certain images' nuances or cultural sensitivities, leading to inappropriate or offensive outputs.

\section{Artifacts}
All datasets and models are publicly accessible for academic use, and the official OpenAI API is available for academic purposes.

\section{Computational Experiments}
All open-source model experiments were conducted on an NVIDIA RTX A6000 GPU. Approximately 500 US dollars were spent on an OpenAI API call for GPT-4v and GPT-4o experiments.

\section{Acknowledgement}
We thank Kai Zhang, Yujie Lu, and Tianle Li for the discussion. We also thank Xueguang Ma and Tianle Li for sharing the quotas of GPT-4v and GPT-4o API calls from OpenAI.

\bibliography{acl_latex}
\clearpage
\appendix

\section{Prompt Templates}
\label{sec:appendix_prompt}

\textbf{Prompt Engineering.} We found that not all MLLMs can fully understand our prompt to give a desired output format consistently. Thus we required MLLMs to output a JSON format, which is supposed to be capable for most MLLMs. 

\textbf{Prompt Design.} The prompt is divided into two segments: the `context prompt' and the `rating prompt'. The ultimate prompt provided to the model is a combination of these two segments.

\NewTColorBox{Context_Box}{ s O{!htbp} }{%
  floatplacement={#2},
  IfBooleanTF={#1}{float*,width=\textwidth}{float},
  colframe=gray!50!black,colback=gray!10!white,title=Context,
  }

\NewTColorBox{PQ_ALL_Box}{ s O{!htbp} }{%
  floatplacement={#2},
  IfBooleanTF={#1}{float*,width=\textwidth}{float},
  colframe=cyan!50!black,colback=cyan!10!white,title=PQ Rating Prompt Template (for all tasks),
  }

\NewTColorBox{SC_TIE_Box}{ s O{!htbp} }{%
  floatplacement={#2},
  IfBooleanTF={#1}{float*,width=\textwidth}{float},
  colframe=yellow!50!black,colback=yellow!10!white,title=SC Rating Prompt Template (Text/Mask-Guided Image Editing)
  }

\NewTColorBox{SC_T2I_Box}{ s O{!htbp} }{%
  floatplacement={#2},
  IfBooleanTF={#1}{float*,width=\textwidth}{float},
  colframe=pink!50!black,colback=pink!10!white,title=SC Rating Prompt Template (Text-Guided Image Generation)
  }

\NewTColorBox{SC_CIG_Box}{ s O{!htbp} }{%
  floatplacement={#2},
  IfBooleanTF={#1}{float*,width=\textwidth}{float},
  colframe=green!50!black,colback=green!10!white,title=SC Rating Prompt Template (Control-Guided Image Generation)
  }

\NewTColorBox{SC_SDIG_Box}{ s O{!htbp} }{%
  floatplacement={#2},
  IfBooleanTF={#1}{float*,width=\textwidth}{float},
  colframe=orange!50!black,colback=orange!10!white,title=SC Rating Prompt Template (Subject-Driven Image Generation)
  }

\NewTColorBox{SC_SDIE_Box}{ s O{!htbp} }{%
  floatplacement={#2},
  IfBooleanTF={#1}{float*,width=\textwidth}{float},
  colframe=magenta!50!black,colback=magenta!10!white,title=SC Rating Prompt Template (Subject-Guided Image Editing)
  }

\begin{Context_Box}[!ht]
You are a professional digital artist. You will have to evaluate the effectiveness of the AI-generated image(s) based on the given rules.
You will have to give your output in this way (Keep your reasoning concise and short.):\\
\{\\
"score" : [...],\\
"reasoning" : "..."\\
\}\\
\end{Context_Box}

\begin{PQ_ALL_Box}[!ht]
 
RULES:

The image is an AI-generated image.
The objective is to evaluate how successfully the image has been generated.

On a scale 0 to 10: \\
A score from 0 to 10 will be given based on image naturalness. \\
( 0 indicates that the scene in the image does not look natural at all or gives an unnatural feeling such as a wrong sense of distance, wrong shadow, or wrong lighting. 10 indicates that the image looks natural. )\\
A second score from 0 to 10 will rate the image artifacts. \\
(
    0 indicates that the image contains a large portion of distortion, watermarks, scratches, blurred faces, unusual body parts, or subjects not harmonized. 
    10 indicates the image has no artifacts.
)\\
Put the score in a list such that output score = [naturalness, artifacts]

\end{PQ_ALL_Box}

\begin{SC_T2I_Box}*[!ht]
 
RULES:\\
The image is an AI-generated image according to the text prompt.
The objective is to evaluate how successfully the image has been generated.\\
On a scale 0 to 10: \\
A score from 0 to 10 will be given based on the success in following the prompt. 
(0 indicates that the AI-generated image does not follow the prompt at all. 10 indicates the AI-generated image follows the prompt perfectly.)\\
Put the score in a list such that output score = [score].\\
Text Prompt: <prompt>
\end{SC_T2I_Box}

\begin{SC_TIE_Box}*[!ht]
 
RULES: \\
Two images will be provided: The first being the original AI-generated image and the second being an edited version of the first.
The objective is to evaluate how successfully the editing instruction has been executed in the second image.
Note that sometimes the two images might look identical due to the failure of the image edit.\\
On scale of 0 to 10: \\
A score from 0 to 10 will be given based on the success of the editing.
(0 indicates that the scene in the edited image does not follow the editing instructions at all. 10 indicates that the scene in the edited image follows the editing instruction text perfectly.) \\
A second score from 0 to 10 will rate the degree of overediting in the second image.
(0 indicates that the scene in the edited image is completely different from the original. 10 indicates that the edited image can be recognized as a minimally edited yet effective version of the original.)\\
Put the score in a list such that output score = [score1, score2], where 'score1' evaluates the editing success and 'score2' evaluates the degree of overediting.\\
Editing instruction: <instruction>
\end{SC_TIE_Box}

\begin{SC_CIG_Box}*[!ht]
 
RULES:\\
Two images will be provided: The first being a processed image (e.g. Canny edges, openpose, grayscale, etc.) and the second being an AI-generated image using the first image as guidance. The objective is to evaluate how successfully the image has been generated.\\
On scale 0 to 10: \\
A score from 0 to 10 will be given based on the success in following the prompt. 
(0 indicates that the second image does not follow the prompt at all. 10 indicates the second image follows the prompt perfectly.)\\
A second score from 0 to 10 will rate how well the generated image is following the guidance image. 
(0 indicates that the second image does not follow the guidance at all. 10 indicates that the second image is following the guidance image.)\\
Put the score in a list such that output score = [score1, score2], where 'score1' evaluates the prompt and 'score2' evaluates the guidance.\\
Text Prompt: <prompt>\\
\end{SC_CIG_Box}

\begin{SC_SDIG_Box}*[!ht]
 
RULES:\\
Two images will be provided: The first is a token subject image and the second is an AI-generated image using the first image as guidance.
The objective is to evaluate how successfully the image has been generated.\\
On a scale of 0 to 10: \\
A score from 0 to 10 will be given based on the success in following the prompt. 
(0 indicates that the second image does not follow the prompt at all. 10 indicates the second image follows the prompt perfectly.)\\
A second score from 0 to 10 will rate how well the subject in the generated image resembles the token subject in the first image. 
(0 indicates that the subject in the second image does not look like the token subject at all. 10 indicates the subject in the second image looks exactly like the token subject.)\\
Put the score in a list such that output score = [score1, score2], where 'score1' evaluates the prompt and 'score2' evaluates the resemblance.\\
Text Prompt: <prompt>
\end{SC_SDIG_Box}

\begin{SC_SDIE_Box}*[!ht]
 
RULES:\\
Three images will be provided: 
The first image is an input image to be edited.
The second image is a token subject image.
The third image is an AI-edited image from the first image. it should contain a subject that looks like the subject in the second image.
The objective is to evaluate how successfully the image has been edited.\\
On a scale 0 to 10: \\
A score from 0 to 10 will rate how well the subject in the generated image resembles the token subject in the second image. 
(0 indicates that the subject in the third image does not look like the token subject at all. 10 indicates the subject in the third image looks exactly like the token subject.)\\
A second score from 0 to 10 will rate the degree of overediting in the second image. 
(0 indicates that the scene in the edited image is completely different from the first image. 10 indicates that the edited image can be recognized as a minimally edited yet effective version of the original.)\\
Put the score in a list such that output score = [score1, score2], where 'score1' evaluates the resemblance and 'score2' evaluates the degree of overediting.\\
Subject: <subject>
\end{SC_SDIE_Box}

\clearpage
\section{Supplementary Information}
\label{sec:supp}

\subsection{Human Correlation Study}

In our paper content, we only reported the Spearman correlations. Here we included the Pearson and Kendall correlation
in Table \ref{tab:app_humanhuman_corr} for comparative analysis of Human-to-Human (H-H) correlation.

\subsection{Zero-shot vs One-shot on VIE}

We applied only zero-shot and one-shot experiments in this paper because not even GPT-4v can produce anything with few-shot setting in our context. We report full table of GPT-4v performance in Table \ref{tab:app_GPT-4v_human_corr} for zero-shot vs one-shot results. 

\subsection{Autometrics vs Human Detail results}

See Table \ref{tab:clipScore_corr}, \ref{tab:lpips_corr}, \ref{tab:dino_corr}, \ref{tab:CLIP-I_corr} for detail statistics of CLIP-Score, LPIPS, DINO, and CLIP-I correlation with human ratings.

\subsection{ImagenHub Models used}

\begin{itemize}
  \setlength\itemsep{0em}
  \item Text-guided Image Generation: Stable Diffusion (SD) \citep{rombach2022high}, SDXL \citep{sd-xl}, DALLE-2 \citep{ramesh2022hierarchical}, DeepFloydIF \citep{deep-floyd}, OpenJourney \citep{openjourney}.
  \item Mask-guided Image Editing: SD \citep{sdinpaint}, SDXL \citep{sd-xl}, GLIDE, BlendedDiffusion \citep{avrahami2022blended}
  \item Text-guided Image Editing: MagicBrush \citep{zhang2023magicbrush}, InstructPix2Pix \citep{brooks2022instructpix2pix}, Prompt-to-Prompt \citep{mokady2023null}, CycleDiffusion \citep{cyclediffusion}, SDEdit \citep{meng2021sdedit}, Text2Live \citep{bar2022text2live}, DiffEdit \citep{couairon2022diffedit}, Pix2PixZero \citep{parmar2023zero}.
  \item Subject-driven Image Generation: DreamBooth \citep{ruiz2023dreambooth}, DreamBooth-Lora \citep{hu2021lora}, BLIP-Diffusion \citep{li2023blip}, TextualInversion \citep{gal2022image}.
  \item Subject-driven Image Editing: PhotoSwap \citep{gu2023photoswap}, DreamEdit \citep{li2023dreamedit}, BLIP-Diffusion.
  \item Multi-concept Image Composition: CustomDiffusion \citep{kumari2023multi}, DreamBooth, TextualInversion.
  \item Control-guided Image Generation: ControlNet \citep{zhang2023adding}, UniControl \citep{qin2023unicontrol}.
\end{itemize}

\subsection{ImagenHub Human data information}

We showed the total human rating data we used for each task in Table \ref{tab:imagenhub_data_info}.

\begin{table*}[!ht]
\centering
\scalebox{0.85}{
\begin{tabular}{l|ccc|ccc}
\toprule
& \multicolumn{6}{c}{Backbone: GPT-4v} \\
& \multicolumn{3}{c}{Zero-Shot} & \multicolumn{3}{c}{One-Shot} \\
Image Model & M-H$^{SC}_{\text{0shot}}$ & M-H$^{PQ}_{\text{0shot}}$ & M-H$^{O}_{\text{0shot}}$ & M-H$^{SC}_{\text{1shot}}$ & M-H$^{PQ}_{\text{1shot}}$ & M-H$^{O}_{\text{1shot}}$ \\
\midrule
\multicolumn{7}{c}{Text-guided Image Generation} \\
\midrule
DeepFloydIF & 0.5182 & 0.3509 & 0.5479 & 0.4272 & 0.2048 & 0.3849\\
Stable Diffusion XL & 0.5684 & 0.2823 & 0.5301 & 0.5136 & 0.1522 & 0.3735\\
Dalle-2 & 0.5046 & 0.2192 & 0.4871 & 0.4469 & 0.1822 & 0.5364\\
OpenJourney& 0.4835 & 0.1624 & 0.4648 & 0.4563 & 0.2730 & 0.3750\\
Stable Diffusion 2.1 & 0.5957 & 0.1981 & 0.4658 & 0.5988 & 0.0820 & 0.3311\\
\midrule
\multicolumn{7}{c}{Mask-guided Image Editing} \\
\midrule
SDXL-Inpainting & 0.5461 & 0.2331 & 0.4772 & 0.5308 & 0.3460 & 0.5261\\
SD-Inpainting & 0.5607 & 0.4253 & 0.544 & 0.3759 & 0.3446 & 0.3969\\
GLIDE & 0.4663 & 0.2816 & 0.4499 & 0.4247 & 0.1056 & 0.3536\\
BlendedDiffusion & 0.3695 & 0.2363 & 0.2563 & 0.4054 & 0.1624 & 0.3283\\
\midrule
\multicolumn{7}{c}{Text-guided Image Editing} \\
\midrule
MagicBrush        & 0.3273 & 0.3696 & 0.3395 & 0.3613 & 0.5135 & 0.4727\\
InstructPix2Pix   & 0.3094 & 0.4461 & 0.3363 & 0.4423 & 0.3106 & 0.3921\\
Prompt-to-prompt  & 0.3094 & 0.3696 & 0.3395 & 0.2514 & 0.2057 & 0.2068\\
CycleDiffusion    & 0.4488 & 0.6124 & 0.3927 & 0.3522 & 0.3374 & 0.1578\\
SDEdit            & 0.1607 & 0.3944 & 0.1570 & 0.1754 & 0.3837 & 0.2814\\
Text2Live         & 0.1875 & 0.4158 & 0.1964 & 0.2817 & 0.2357 & 0.2753\\
DiffEdit          & 0.1803 & 0.5957 & 0.0247 & 0.1761 & 0.4874 & 0.1281\\
Pix2PixZero       & 0.2144 & 0.4502 & 0.2193 & -0.0588 & 0.3609 & -0.0588\\
\midrule
\multicolumn{7}{c}{Subject-driven Image Generation} \\
\midrule
DreamBooth        & 0.4975 & 0.2199 & 0.4787 & 0.5409 & 0.1930 & 0.5848\\
BLIP-Diffusion    & 0.3367 & 0.0663 & 0.2845 & 0.1176 & 0.3402 & 0.1194\\
TextualInversion  & 0.5564 & 0.2398 & 0.4795 & 0.3882 & 0.0010 & 0.3035\\
DreamBooth-Lora   & 0.2938 & 0.2448 & 0.3285 & 0.0856 & 0.3860 & 0.1225\\
\midrule
\multicolumn{7}{c}{Subject-driven Image Editing} \\
\midrule
PhotoSwap         & 0.3711 & 0.1246 & 0.1598 & -0.0782 & 0.0385 & -0.1063\\
DreamEdit         & 0.3817 & 0.4419 & 0.1580 & 0.1384 & 0.3037 & 0.0954\\
BLIP-Diffusion    & 0.2671 & 0.3488 & 0.1379 & -0.1368 & 0.1333 & -0.0309\\
\midrule
\multicolumn{7}{c}{Multi-concept Image Composition} \\
\midrule
CustomDiffusion   & 0.4781 & 0.431 & 0.4263 & 0.5064 & 0.0194 & 0.4867\\
DreamBooth        & 0.1494 & 0.2367 & 0.232 & 0.0396 & 0.0633 & 0.0694\\
TextualInversion  & 0.3703 & 0.269 & 0.3857 & 0.0183 & 0.2745 & 0.0266\\
\midrule
\multicolumn{7}{c}{Control-guided Image Generation} \\
\midrule
ControlNet        & 0.4270 & 0.4827 & 0.4753 & 0.3561 & 0.4052 & 0.4055\\
UniControl        & 0.5797 & 0.4173 & 0.3972 & 0.4655 & 0.4737 & 0.4988\\
\bottomrule
\end{tabular} }
\caption{Comprehensive study on the Spearman correlation between GPT-4v-to-Human (GPT-4v-H) ratings across various models, in zero-shot (0shot) and one-shot (1shot) settings, across different test categories: Semantic Consistency (SC), Perceptual Quality (PQ), and Overall (O). }
\label{tab:app_GPT-4v_human_corr}
\end{table*}

\begin{table*}[!ht]
\centering
\scalebox{0.80}{
\begin{tabular}{l|ccc|ccc|ccc}
\toprule
Image Model & H-H$^{SC}_{\text{pear}}$ & H-H$^{PQ}_{\text{pear}}$ & H-H$^{O}_{\text{pear}}$ & H-H$^{SC}_{\text{spea}}$ & H-H$^{PQ}_{\text{spea}}$ & H-H$^{O}_{\text{spea}}$ & H-H$^{SC}_{\text{kend}}$ & H-H$^{PQ}_{\text{kend}}$ & H-H$^{O}_{\text{kend}}$ \\
\midrule
\multicolumn{10}{c}{Text-guided Image Generation} \\
\midrule
DeepFloydIF          & 0.5933 & 0.3086 & 0.5595 & 0.5635 & 0.3029 & 0.5131 & 0.5360 & 0.2878 & 0.4581 \\
Stable Diffusion XL  & 0.5990 & 0.4957 & 0.5945 & 0.5807 & 0.4992 & 0.5896 & 0.5468 & 0.4719 & 0.5289 \\
Dalle-2              & 0.5208 & 0.5024 & 0.4630 & 0.5019 & 0.4680 & 0.4348 & 0.4654 & 0.4459 & 0.3820 \\
OpenJourney          & 0.5678 & 0.3853 & 0.5513 & 0.5321 & 0.3600 & 0.4861 & 0.5017 & 0.3442 & 0.4347 \\
Stable Diffusion 2.1 & 0.6202 & 0.3227 & 0.5397 & 0.5979 & 0.2772 & 0.4962 & 0.5707 & 0.2636 & 0.4557 \\
\midrule
\multicolumn{10}{c}{Mask-guided Image Editing} \\
\midrule
SDXL-Inpainting  & 0.6550 & 0.5929 & 0.6578 & 0.6574 & 0.5928 & 0.6556 & 0.6160 & 0.5382 & 0.6040 \\
SD-Inpainting    & 0.6606 & 0.5197 & 0.5716 & 0.6590 & 0.5166 & 0.5394 & 0.6222 & 0.4728 & 0.5039 \\
GLIDE            & 0.6253 & 0.5496 & 0.6144 & 0.5894 & 0.5530 & 0.5695 & 0.5573 & 0.4984 & 0.5357 \\
BlendedDiffusion & 0.5863 & 0.5873 & 0.5879 & 0.5051 & 0.5511 & 0.4224 & 0.4911 & 0.5346 & 0.4157 \\
\midrule
\multicolumn{10}{c}{Text-guided Image Editing} \\
\midrule
MagicBrush        & 0.6217 & 0.5251 & 0.6288 & 0.6219 & 0.5190 & 0.6289 & 0.5740 & 0.4740 & 0.5651 \\
InstructPix2Pix   & 0.6573 & 0.6158 & 0.6632 & 0.6600 & 0.5955 & 0.6561 & 0.6250 & 0.5502 & 0.6157 \\
Prompt-to-prompt  & 0.5954 & 0.5084 & 0.5699 & 0.5880 & 0.5028 & 0.5811 & 0.5611 & 0.4537 & 0.5470 \\
CycleDiffusion    & 0.5908 & 0.5848 & 0.6101 & 0.5482 & 0.5887 & 0.5891 & 0.5228 & 0.5378 & 0.5600 \\
SDEdit            & 0.2303 & 0.4717 & 0.1674 & 0.2657 & 0.4705 & 0.1991 & 0.2618 & 0.4211 & 0.1957 \\
Text2Live         & 0.3167 & 0.6013 & 0.2890 & 0.2675 & 0.5757 & 0.1524 & 0.2648 & 0.5440 & 0.1503 \\
DiffEdit          & 0.2513 & 0.6331 & 0.3570 & 0.3286 & 0.6214 & 0.4265 & 0.3268 & 0.5924 & 0.4247 \\
Pix2PixZero       & 0.4747 & 0.5763 & 0.5247 & 0.3311 & 0.5770 & 0.3327 & 0.3305 & 0.5299 & 0.3312 \\
\midrule
\multicolumn{10}{c}{Subject-driven Image Generation} \\
\midrule
DreamBooth        & 0.6337 & 0.3988 & 0.5834 & 0.6452 & 0.3871 & 0.6208 & 0.6010 & 0.3787 & 0.5625 \\
BLIP-Diffusion    & 0.4970 & 0.2663 & 0.4394 & 0.4458 & 0.3263 & 0.4390 & 0.4090 & 0.3180 & 0.3993 \\
TextualInversion  & 0.5987 & 0.3219 & 0.5533 & 0.6000 & 0.3351 & 0.5686 & 0.5683 & 0.3078 & 0.5226 \\
DreamBooth-Lora   & 0.5014 & 0.4571 & 0.4278 & 0.3903 & 0.4430 & 0.3878 & 0.3831 & 0.4169 & 0.3756 \\
\midrule
\multicolumn{10}{c}{Subject-driven Image Editing} \\
\midrule
PhotoSwap         & 0.4685 & 0.3213 & 0.5025 & 0.4805 & 0.2961 & 0.4973 & 0.4412 & 0.2768 & 0.4368 \\
DreamEdit         & 0.5684 & 0.2319 & 0.5520 & 0.5867 & 0.2245 & 0.5460 & 0.5485 & 0.2130 & 0.4892 \\
BLIP-Diffusion    & 0.5411 & 0.4086 & 0.5074 & 0.5359 & 0.4033 & 0.5051 & 0.5221 & 0.3779 & 0.4857 \\
\midrule
\multicolumn{10}{c}{Multi-concept Image Composition} \\
\midrule
CustomDiffusion   & 0.7257 & 0.4889 & 0.7215 & 0.7256 & 0.4838 & 0.7217 & 0.7101 & 0.4665 & 0.6963 \\
DreamBooth        & 0.6560 & 0.6583 & 0.6575 & 0.6209 & 0.6423 & 0.6222 & 0.6068 & 0.6228 & 0.6022 \\
TextualInversion  & 0.6833 & 0.6009 & 0.6799 & 0.6990 & 0.5803 & 0.6980 & 0.6935 & 0.5563 & 0.6898 \\
\midrule
\multicolumn{10}{c}{Control-guided Image Generation} \\
\midrule
ControlNet        & 0.6166 & 0.5730 & 0.5830 & 0.6144 & 0.5682 & 0.5868 & 0.5585 & 0.5408 & 0.5429 \\
UniControl        & 0.6014 & 0.6194 & 0.6131 & 0.6060 & 0.6062 & 0.5954 & 0.5577 & 0.5741 & 0.5533 \\
\bottomrule
\end{tabular} }
\caption{Comparative analysis of Human-to-Human (H-H) correlation ratings across various models. Metrics used include Pearson's (pear), Spearman's (spea), and Kendall's (kend) correlation coefficients, across different test categories: Semantic Consistency (SC), Perceptual Quality (PQ), and Overall (O).}
\label{tab:app_humanhuman_corr}
\end{table*}

\begin{table}[!ht]
\centering
\scalebox{0.85}{
\begin{tabular}{l|ccc}
\toprule
& \multicolumn{3}{c}{Metric: CLIP-Score} \\
Image Model & M-H$^{SC}_{\text{corr}}$ & M-H$^{PQ}_{\text{corr}}$ & M-H$^{O}_{\text{corr}}$ \\
\midrule
\multicolumn{4}{c}{Text-guided Image Generation} \\
\midrule
DeepFloydIF  & 0.0272 & 0.1025 & 0.0332 \\
OpenJourney  & -0.1628 & -0.0875 & -0.1907 \\
DALLE2       & -0.0946 & 0.1469 & -0.0381 \\
SD           & -0.0528 & -0.0691 & -0.0632 \\
SDXL         & -0.1265 & -0.1500 & -0.1830 \\
\bottomrule
\end{tabular} }
\caption{CLIP-Score vs Human correlation on Text-guided image generation task.}
\label{tab:clipScore_corr}
\end{table}

\begin{table}[!ht]
\centering
\scalebox{0.85}{
\begin{tabular}{l|ccc}
\toprule
& \multicolumn{3}{c}{Metric: LPIPS} \\
Image Model & M-H$^{SC}_{\text{corr}}$ & M-H$^{PQ}_{\text{corr}}$ & M-H$^{O}_{\text{corr}}$ \\
\midrule
\multicolumn{4}{c}{Text-guided Image Editing} \\
\midrule
InstructPix2Pix   & 0.1652 & 0.4717 & 0.2045 \\
CycleDiffusion    & -0.0936 & 0.3193 & -0.0211 \\
MagicBrush        & 0.2146 & 0.3722 & 0.2667 \\
Text2Live         & -0.0812 & 0.2906 & -0.0787 \\
DiffEdit          & 0.0943 & 0.3299 & 0.1440 \\
Pix2PixZero       & 0.1379 & 0.0256 & 0.1370 \\
Prompt2prompt     & 0.1918 & 0.1929 & 0.1798 \\
SDEdit            & 0.1381 & 0.0441 & 0.0857 \\
\midrule
\multicolumn{4}{c}{Mask-guided Image Editing} \\
\midrule
Glide             & -0.1098 & 0.0647 & -0.0662 \\
BlendedDiffusion  & 0.0980 & 0.1371 & 0.0598 \\
SDInpaint         & -0.2447 & -0.0749 & -0.2110 \\
SDXLInpaint       & -0.1496 & 0.1318 & -0.0607 \\
\midrule
\multicolumn{4}{c}{Control-guided Image Generation} \\
\midrule
ControlNet         & 0.3447 & 0.3916 & 0.3888 \\
UniControl       & 0.4319 & 0.5048 & 0.4904 \\
\bottomrule
\end{tabular} }
\caption{LPIPS (signs inverted) vs Human correlation on several tasks.}
\label{tab:lpips_corr}
\end{table}

\begin{table}[!ht]
\centering
\scalebox{0.85}{
\begin{tabular}{l|ccc}
\toprule
& \multicolumn{3}{c}{Metric: DINO} \\
Model & M-H$^{SC}_{\text{corr}}$ & M-H$^{PQ}_{\text{corr}}$ & M-H$^{O}_{\text{corr}}$ \\
\midrule
\multicolumn{4}{c}{Multi-Concept Image Composition} \\
\midrule
TextualInversion   & 0.0759 & -0.2746 & 0.0754 \\
DreamBooth         & 0.1027 & -0.0761 & 0.1054 \\
CustomDiffusion    & 0.1159 & -0.1466 & 0.1074 \\
\midrule
\multicolumn{4}{c}{Subject-Driven Image Generation} \\
\midrule
DreamBoothLora     & 0.2335 & 0.0684 & 0.2535 \\
BLIPDiffusion (Gen)  & 0.4718 & 0.0798 & 0.4751 \\
TextualInversion   & 0.6508 & -0.0169 & 0.6450 \\
DreamBooth         & 0.4153 & 0.3535 & 0.4396 \\
\midrule
\multicolumn{4}{c}{Subject-Driven Image Editing} \\
\midrule
BLIPDiffusion (Edit)  & 0.4063 & -0.1081 & 0.4000 \\
DreamEdit           & 0.1994 & -0.0877 & 0.1878 \\
PhotoSwap           & 0.3300 & 0.0814 & 0.3424 \\
\bottomrule
\end{tabular}
}
\caption{DINO vs Human correlation on several tasks.}
\label{tab:dino_corr}
\end{table}

\begin{table}[!ht]
\centering
\scalebox{0.85}{
\begin{tabular}{l|ccc}
\toprule
& \multicolumn{3}{c}{Metric: CLIP-I} \\
Image Model & M-H$^{SC}_{\text{corr}}$ & M-H$^{PQ}_{\text{corr}}$ & M-H$^{O}_{\text{corr}}$ \\
\midrule
\multicolumn{4}{c}{Multi-Concept Image Composition} \\
\midrule
TextualInversion   & 0.1511 & -0.1847 & 0.1523 \\
DreamBooth         & 0.0741 & -0.1166 & 0.0758 \\
CustomDiffusion    & 0.2319 & 0.0116 & 0.2246 \\
\midrule
\multicolumn{4}{c}{Subject-Driven Image Generation} \\
\midrule
DreamBoothLora       & 0.2499 & 0.1801 & 0.2615 \\
BLIPDiffusion (Gen)    & 0.2611 & 0.1031 & 0.2660 \\
TextualInversion     & 0.5776 & 0.0362 & 0.5775 \\
DreamBooth           & 0.1324 & 0.3648 & 0.1587 \\
\midrule
\multicolumn{4}{c}{Subject-Driven Image Editing} \\
\midrule
BLIPDiffusion (Edit)  & 0.4202 & -0.0798 & 0.3844 \\
DreamEdit           & 0.1927 & 0.1535 & 0.1955 \\
PhotoSwap           & 0.2613 & 0.3028 & 0.2874 \\

\bottomrule
\end{tabular} }
\caption{CLIP-I vs Human correlation on several tasks.}
\label{tab:CLIP-I_corr}
\end{table}

\begin{table}[!ht]
\centering
\scalebox{0.85}{
\begin{tabular}{c|c}
\toprule
Data amount per model & Total Human rating data  \\
\midrule
\multicolumn{2}{c}{Task: Text-guided Image Generation} \\
\midrule
197 & 2955 \\
\midrule
\multicolumn{2}{c}{Task: Mask-guided Image Editing} \\
\midrule
179 & 2148 \\
\midrule
\multicolumn{2}{c}{Task: Text-guided Image Editing} \\
\midrule
179 & 4296 \\
\midrule
\multicolumn{2}{c}{Task: Subject-driven Image Generation} \\
\midrule
150 & 1800 \\
\midrule
\multicolumn{2}{c}{Task: Subject-driven Image Editing} \\
\midrule
154 & 1386 \\
\midrule
\multicolumn{2}{c}{Task: Multi-concept Image Composition} \\
\midrule
102 & 918 \\
\midrule
\multicolumn{2}{c}{Task: Control-guided Image Generation} \\
\midrule
150 & 900 \\
\midrule
& Sum of 7 tasks \\
\midrule
& 14403 \\
\bottomrule

\end{tabular} }
\caption{Number of human ratings from ImagenHub used in this paper.}
\label{tab:imagenhub_data_info}
\end{table}

\clearpage
\section{Backbone Performances}
\label{sec:backbone_performance}

\subsection{Parsing MLLM outputs}
We tried to parse the output using Regex and modify the format requirement if the MLLM fail to do so. If the output failed to pass our parsing rules, we fill random value as output to penalize the correlation.

\subsection{Observations of GPT-4o.}
GPT-4o-2024-05-13 tends to be the best MLLM in this context. It can understand every task instruction in \score and produce reasonable scores and rationale.

\subsection{Observations of Gemini-Pro.}
Gemini-1.5-Pro can understand every task instruction in \score and produce reasonable scores and rationale, achieving similar performance as GPT-4v.

\subsection{Observations of GPT-4v.}
GPT-4-vision-preview tends to be the second best MLLM in this context. It can understand every task instruction in \score and produce reasonable scores and rationale.

\subsection{Observations of LLaVA.}
LLaVA-1.5-7B can also understand every task instruction in \score and produce reasonable rationale. However, the scores produced tend to be concentrated toward certain numbers.

\subsection{Observations of Qwen-VL.}
Qwen-VL-7B does not understand the meaning of delimiter. However, it was able to output a JSON-like dictionary following the instructions on both SC and PQ. The rationale produced is often not reasonable.

\subsection{Observations of BLIP2.}
BLIP-2 FLAN-T5-XXL often failed to produce the result formats according to the instructions, especially in PQ. It also tend to give 0 score in SC. Prompt engineering in our context does not solve the issue.

\subsection{Observations of InstructBLIP.}
InstructBLIP-T5-XL shares same observation as BLIP-2 FLAN-T5-XXL. 

\subsection{Observations of Fuyu.}
Fuyu-8B always output 0 and failed to follow in our instruction. 

\subsection{Observations of CogVLM.}
CogVLM tends to output numbers fall off the range [0, 10] and often failed to follow the required format. Prompt engineering in our context does not solve the issue.

\subsection{Observations of OpenFlamingo.}
OpenFlamingo simply printing blank as the output in our context. 

\end{document}